\documentclass{article}
\usepackage{iclr2027_conference,times}

\usepackage{amsmath,amsfonts,bm}

\def\eqref#1{equation~\ref{#1}}

\def\1{\bm{1}}

\def\mB{{\bm{B}}}

\DeclareMathAlphabet{\mathsfit}{\encodingdefault}{\sfdefault}{m}{sl}
\SetMathAlphabet{\mathsfit}{bold}{\encodingdefault}{\sfdefault}{bx}{n}

\def\sC{{\mathbb{C}}}

\def\sK{{\mathbb{K}}}
\def\sL{{\mathbb{L}}}

\def\sP{{\mathbb{P}}}

\def\sS{{\mathbb{S}}}

\usepackage{hyperref}
\usepackage{url}
\usepackage{cleveref}
\usepackage{amsmath}
\usepackage{amssymb}
\usepackage{stmaryrd}
\usepackage{graphicx}
\usepackage{booktabs}
\usepackage{multirow}
\usepackage{placeins}
\usepackage{wrapfig}

\title{Sparse cubical complexes for efficient topology-preservation in image data}

\author{Alexander H.~Berger$^{1,3}$, Marco Fontana$^{3}$, Daniel Rueckert$^{3,4,5}$, Johannes C.~Paetzold$^{1,2}$, \\
\bf Laurin Lux$^{1,3,\dagger}$, Ulrich Bauer$^{3,5,6,\dagger}$ \\
\normalfont $^{1}$Weill Cornell Medicine, New York, USA \\
\normalfont $^{2}$Cornell Tech, New York, USA \\
\normalfont $^{3}$Technical University of Munich, Munich, Germany \\
\normalfont $^{4}$Department of Computing, Imperial College London, UK \\
\normalfont $^{5}$Munich Center for Machine Learning (MCML), Munich, Germany \\
\normalfont $^{6}$Munich Data Science Institute, Technical University of Munich, Munich, Germany
}

\iclrfinalcopy

\begin{document}

\maketitle
\lhead{Preprint}
{\renewcommand{\thefootnote}{}\footnotetext{Corresponding author: \texttt{a.berger@tum.de} \qquad $^{\dagger}$These authors contributed equally as senior authors}}

\begin{abstract}
Persistent homology (PH) is a frequently used tool for extracting and preserving topological information from image data, particularly in image segmentation, where preservation of topological structures is important. However, despite its general applicability across dimensionality, domains, and target structures, the runtime cost of PH-based methods often makes their practical use infeasible. In this work, we argue that this runtime cost is largely driven by processing information that is unimportant for downstream application (e.g. as optimization objective). We propose sparse cubical filtrations as an alternative foundation for PH computation, reducing subsequent computational costs by factors of up to 100 on real datasets. We show close agreement with the optimization signal of the dense counterpart and empirically evaluate our solution's effectiveness as an optimization objective in realistic training regimes where other PH-based objectives can practically not operate (i.e., 3D data with large patch sizes). We show how our solution improves topological accuracy by up to 80\% across six diverse datasets while maintaining pixel- and region-based accuracy.

\end{abstract}

\section{Introduction}
\label{sec:intro}
Persistent homology (PH) is one of the most widely used tools for describing the topology of an input space. It has wide applications in machine learning, particularly in computer vision, where it has been used for image reconstruction \citep{moor2020topological}, generation \citep{gupta2025topodiffusionnet}, and frequently, segmentation \citep{stucki2023topologically,stucki2024efficient,berger2024topologically,berger2025pitfalls,hu2019topology,hu2021topology,hu2022structure,clough2020topological,qi2023dynamic,byrne2022persistent}. This development is driven by widespread demand for topologically accurate segmentations in downstream applications in domains that include connectomics \citep{funke2018large}, flow simulations \citep{alastruey2007modelling}, and histopathology \citep{xu2025topocellgen}.

Using PH on digital images involves building cubical complexes, in which each voxel is represented by cells of the complex. These cubical complexes serve as the basis for subsequent computations to obtain persistence diagrams or persistence barcodes (see \cref{fig:barcode_example}). A persistence barcode describes an image's topology at every possible binarization threshold and is then processed for the aforementioned use cases, e.g., as an optimization objective in image segmentation. PH-based methods are appealing because they are general, i.e., they are not restricted to specific input dimensions or tailored to specific target structures (e.g., tubular structures), and they can be tuned towards reducing a specific error type (e.g., spurious components or false splits).

\begin{figure}[t]
    \centering
    \includegraphics[width=\linewidth]{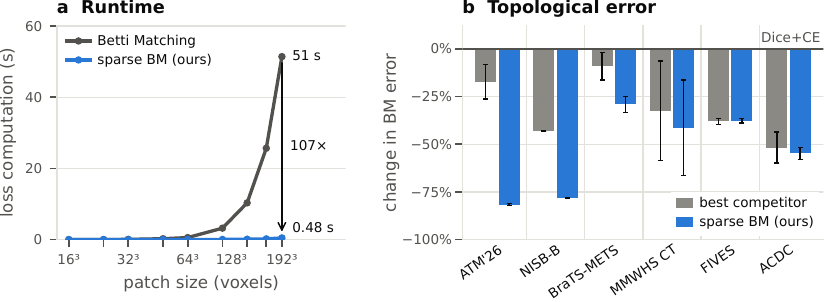}
    \caption{Our approach, PH-based loss functions with sparse cubical complexes, significantly reduces topological errors in image segmentation at a fraction of the cost, making PH-based loss functions usable with large patch sizes and state-of-the-art training paradigms. (a) In realistic 3D training settings (here, on ATM'26), sparseBM is orders of magnitude faster than classical PH-based loss functions and its runtime is substantially less sensitive to patch size in the sparse regime considered here. (b) sparseBM significantly reduces topological error across domains with diverse topological targets and across 2D and 3D (lower is better). }
    \label{fig:fig1}
\end{figure}

Despite their desirable theoretical properties and impressive results, the widespread practical application of PH-based approaches in computer vision has lagged behind, partly due to their high computational cost. PH computation on 3D input has a cubic worst-case complexity in the number of cells (and therefore voxels). Although modern packages \citep{stucki2024efficient, breton2026fast} make the computation vastly more efficient, PH computation on common training patch sizes still lies in the order of seconds (see \cref{fig:fig1}.a), making large-scale training infeasible.

To mitigate the runtime issue of PH-based methods, we propose building \emph{sparse cubical complexes}, i.e., cubical complexes that represent only a subset of the full image. For the prominent application of training segmentation networks, we construct sparse complexes by omitting cells representing confident background regions and use them as the basis of our optimization objective, \textbf{sparseBM}. This approach, for the first time, enables the general use of PH-based loss functions with large patch sizes and state-of-the-art training paradigms.

The efficiency and effectiveness of this approach rest on two assumptions: (1) most datasets and network predictions are sparse in the foreground (making our method efficient, see \cref{fig:fig1}.a), and (2) confident background regions contain no information that is necessary for computing topology-preserving loss functions (making our method effective, see \cref{fig:fig1}.b).

In the following sections, we explain the background of PH computation on images, present how it (and alternative methods) have been used in the literature to obtain topologically accurate segmentation networks, and then present our method and our empirical evaluation for image segmentation.

\section{Background: Images as filtered cubical complexes}
\label{sec:background}

\emph{Persistent homology (PH)} is computed on a filtration of a combinatorial representation of the image. We represent a digital image $I \in [0,1]^{H \times W \times D}$ as a cubical grid complex $\sK$ using the V-construction, where each voxel with value $a$ becomes a vertex of $\sK$ carrying the same value $a$, and every higher-dimensional cell (edges, squares, and cubes in a 3D image) is assigned the maximal value of the vertices it contains. Cubical grid complexes of this kind (or variants thereof, e.g., \citealp{heiss2017streaming}) underlie most PH-based methods in image segmentation \citep{hu2019topology,clough2019explicit,clough2020topological,stucki2023topologically,stucki2024efficient}. Following prior literature, we adopt the convention that \emph{foreground corresponds to low values}, i.e., in a binary label, foreground voxels carry the value $0$ and background voxels the value $1$.

A \emph{sublevel filtration} captures the topological features of an image across all binarization thresholds $t \in [0,1]$ by building a sequence of subcomplexes $\sK_t \subseteq \sK$. $\sK_t$ contains the cells with values $a \leq t$ and corresponds to $I$ binarized at $t$. As $t$ increases, topological features appear and disappear in $\sK_t$. A feature is \emph{born} at the threshold $b$ at which it first appears (e.g., a component forms, a loop closes, a cavity becomes enclosed). A feature \emph{dies} at the threshold $d$ at which it disappears (e.g., a component merges into an older one, a loop or cavity is filled in). The multiset of intervals $(b, d)$ over all features constitutes the \emph{persistence barcode} $\mB(I)$ (equivalently, the persistence diagram) of the image \citep{edelsbrunner2008persistent}. Figure~\ref{fig:barcode_example} shows an exemplary barcode.

Two properties of barcodes on images are important in the following. First, every birth and death value corresponds to a specific voxel of $I$, which is commonly used for computing (partially) differentiable loss functions \citep{carriere2021optimizing,clough2019explicit,clough2020topological,stucki2023topologically,stucki2024efficient,berger2024topologically,hu2019topology}. Second, a feature that never dies is called \emph{essential}.

\emph{Computing a barcode} requires enumerating, sorting, and processing every cell in $\sK$. For a 3D image with $n$ voxels, the V-construction yields approximately $N=8n$ cells, and the full barcode computation has cubic worst-case complexity in the number of cells. While modern algorithms
\citep{kaji2020cubical,stucki2024efficient,breton2026fast} avoid this worst case in practice, their runtime and memory scale with the size of the complex.

\section{Related Work}
\label{sec:related_work}
\subsection{Persistent homology based loss functions for image segmentation}
\label{sec:related_work_ph}
These barcodes and their matching serve as the foundation for many topology-aware image segmentation methods. \citet{hu2021topology}, for example, extract discrete Morse structures from digital images and prune identified critical structures using persistence barcodes. \citet{clough2020topological} minimize the distance between a continuous probability map's barcode $\mB(P)$ (the network's output) and a topological prior, which is only possible if the target topology is known a priori. \citet{hu2019topology} minimize the Wasserstein distance between $\mB(P)$ and the label's barcode $\mB(L)$, which allows for changing topology but disregards spatial correspondence of topological features in the two images. Later works add spatial information heuristically, through height-function filtrations \citep{oner2023persistent}, spatially weighted Wasserstein matching \citep{wen2025topology}, or overlap-based Hungarian matching of persistence regions, which also serves as a ground-truth-free consistency loss for semi-supervised learning \citep{xu2026match}. Betti Matching \citep{stucki2023topologically} achieves this spatial correspondence in a principled manner by matching the resulting barcodes $\mB(L)$ and $\mB(P)$ via a comparison image $C=min(L,P)$. In Betti Matching, two features are matched if and only if they are carried to the same feature of $C$, following the theory of induced matchings \citep{bauer2014induced}. \citet{berger2024topologically} extended this method to multiclass segmentation problems and \citet{stucki2024efficient} drastically improved its runtime by applying optimizations, such as clearing \citep{chen2011persistent}, implicit matrix reduction \citep{stucki2024efficient}, the use of Union-Find (UF) for 0- and top-dimensional features, and skipping emergent pairs during reduction \citep{bauer2021ripser}.

However, computing the matching still requires persistence computations for $P$, $L$, and
$C$ together with two image-persistence computations for the maps from $P$ and $L$ into $C$. All of these computations' runtimes are governed by the size of the underlying cubical complexes, which grow with the \emph{volume} of the image rather than with the content relevant to segmentation. To limit costs, PH losses are commonly evaluated on small patches \citep{oner2023persistent}, possibly selected around suspected errors \citep{ma2026topology}, thereby disregarding global topology and altering the computed persistence. This observation motivates our method.

\subsection{Other topology-aware loss functions for image segmentation}
Other methods avoid PH and the computation of persistence barcodes altogether. clDice \citep{shit2021cldice} is a method specifically designed to preserve connectivity of tubular structures by computing a differentiable soft-skeleton of $P$ and $L$ and computing an overlap-based loss on these. SkelRecall \citep{kirchhoff2024skeleton} improves clDice's runtime and stability for 3D images by only using the skeleton of $L$, which can be precomputed and does not suffer from 3D artifacts, but improves only the recall side. \citet{menten2023skeletonization} addressed the skeletonization problem in 3D by proposing a kernel-based skeletonization algorithm, which further increases runtime. SCNP \citep{valverde2026towards} avoids explicit topology altogether by penalizing each logit with its worst-classified neighbor. Other works approximate PH using Euler characteristics \citep{li2025topology} or, for 2D inputs, by computing connected components at a limited subset of thresholds \citep{qaiser2019fast}. Topograph \citep{lux2025topograph} addressed the runtime problem of PH-based image segmentation methods by using superpixel graphs for identifying topologically critical regions while theoretically guaranteeing homotopy equivalence for zero loss. However, Topograph relies on Alexander duality between 0- and top-dimensional topological features and is therefore only applicable for 2D images.

While all of these methods address individual problems of PH-based methods, none have demonstrated general applicability across diverse target structures, changing topology, and higher-dimensional images (i.e., 3D), while maintaining computational costs that enable large-scale training at realistic patch sizes.

\subsection{Persistent homology in computer vision outside of image segmentation}
PH-based methods are used in generative modeling to create topologically realistic samples \citep{gupta2025topodiffusionnet,wang2020topogan,xu2025topocellgen}. Other work includes the extraction of the topological signatures using PH for image classification \citep{hofer2017deep,lawson2019persistent}.

\section{Method}
\label{sec:method}

\begin{figure}[t]
    \centering
    \includegraphics[width=0.8\linewidth]{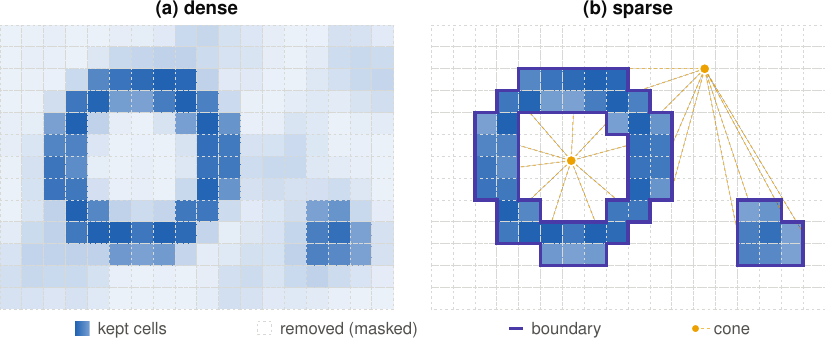}
    \caption{Core idea of the sparse construction. A common retained cubical
    subcomplex $\sS=\sC_\tau$ is selected from the comparison filtration, while
    retained cells keep their original filtration values. The omitted region is
    represented implicitly rather than instantiated as the full cubical grid.}
    \label{fig:core_idea}
\end{figure}

The PH-based methods mentioned in \cref{sec:related_work_ph} operate on the cells of cubical complexes, and their runtime scales with the complexes' sizes (\cref{sec:background}).
We hypothesize that most of the information in these complexes is not required for the downstream application, i.e., in our case, for the computation of a topology-preserving image segmentation loss (or a metric).

Following this rationale, we propose \emph{sparse cubical complexes} as the foundation for computing PH on image data, which accelerates PH computation by restricting
the filtrations to a common cubical subcomplex. More specifically, we propose \textbf{sparseBM}, a topology-preserving loss function following the principles of Betti Matching \citep{stucki2023topologically}. Betti Matching requires persistence computations for the prediction, the label, and a common comparison filtration, together with the corresponding image-persistence computations. We restrict all three filtrations to a common cubical subcomplex selected from the comparison filtration.

\subsection{Sparse cubical filtration}
\label{sec:method_filtrations}

Let $\sK$ be a finite cubical complex and let
$p,\ell,c:\sK\to[0,1]$ denote the prediction, label, and comparison filtration
functions. Write
\[
    \sP_t=\{\sigma:p(\sigma)\leq t\},\qquad
    \sL_t=\{\sigma:\ell(\sigma)\leq t\},\qquad
    \sC_t=\{\sigma:c(\sigma)\leq t\},
\]
and assume
\begin{equation}
    \sP_t\subseteq\sC_t,
    \qquad
    \sL_t\subseteq\sC_t
    \qquad\text{for all }t.
    \label{eq:comparison_inclusions}
\end{equation}
For a threshold $0\leq\tau<1$, define the \emph{retained subcomplex}
\begin{equation}
    \sS:=\sC_\tau.
    \label{eq:retained_complex}
\end{equation}
For $f\in\{p,\ell,c\}$, the sparse filtration is simply the restriction of $f$
to $\sS$,
\begin{equation}
    \sK^{f,\sS}_t:=\sK^f_t\cap\sS,
    \qquad
    \sK^f_t:=\{\sigma:f(\sigma)\leq t\}.
    \label{eq:sparse_restriction}
\end{equation}
\begin{wrapfigure}{r}{0.5\textwidth}
  \centering
  \vspace{15pt}
  \includegraphics[width=0.45\textwidth]{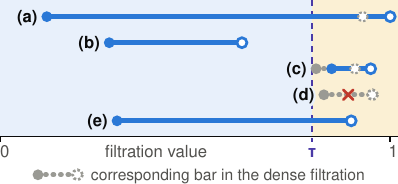}
  \caption{Exemplary comparison between a sparse and dense barcode. Corresponding dense intervals are indicated in grey. The intervals are: (a) kept with shifted death, (b) exact, (c) kept via label with shifted birth and death, (d) omitted, (e) kept exact via label}
  \label{fig:barcode_example}
  \vspace{-15pt}
\end{wrapfigure}
Thus $\tau$ selects which cells are retained but does not truncate their
filtration values. In particular, a cell retained through the comparison
filtration may have prediction or label value larger than $\tau$. Since
$\sP_\tau,\sL_\tau\subseteq\sC_\tau=\sS$, the sparse and dense diagrams
\[
    \sP_t\longrightarrow\sC_t\longleftarrow\sL_t
\]
agree exactly for every $t\leq\tau$.

\subsection{Relation to the dense filtration}
\label{sec:method_dense_sparse}

To compare sparse and dense persistence, extend the restriction to all of
$\sK$ by
\begin{equation}
    \widehat f(\sigma)=
    \begin{cases}
        f(\sigma), & \sigma\in\sS,\\
        1,         & \sigma\notin\sS.
    \end{cases}
    \label{eq:completed_filtration}
\end{equation}
for $f\in\{p,\ell,c\}$. This completion at level $1$ is only a comparison device, and the implementation does not explicitly insert the omitted cells. Because $\sK^f_\tau\subseteq\sS$, every omitted cell satisfies $f(\sigma)>\tau$, so
\begin{equation}
    0\leq\widehat f(\sigma)-f(\sigma)\leq1-\tau.
    \label{eq:filtration_perturbation}
\end{equation}
The inclusions $\sK^{\widehat f}_t\hookrightarrow\sK^f_t$ are isomorphisms through level $\tau$. By the induced matching theorem, they therefore give a specific sparse--dense correspondence in which matched interval endpoints move by at most $1-\tau$, while intervals left unmatched have persistence at most $1-\tau$ \citep{bauer2014induced}. In particular, choosing $\tau$ close to $1$ confines the effect of sparsification to a narrow terminal filtration band, independently of any assumption on the distribution of filtration values.

The same construction applies, through the corresponding commutative squares, to the image-persistence modules associated with
\[
    \sP_\bullet\longrightarrow\sC_\bullet,
    \qquad
    \sL_\bullet\longrightarrow\sC_\bullet.
\]
Thus, the ordinary and image-persistence data entering Betti Matching admit compatible sparse--dense comparisons. This does not imply pairwise equality of the complete sparse and dense Betti matchings above $\tau$, since induced matchings are not functorial in general. Nor can the effect of sparsification in general be described by independently deleting barcode intervals. The effect of sparsification is governed by the kernel, image, and cokernel of the inclusion-induced map, whose persistence was studied by \citet{cohen2009persistent}. Therefore, the inclusion-induced matching is the appropriate comparison.

\subsection{Interpretation for image segmentation}
\label{sec:method_segmentation}

For a binary label and $\tau<1$, the complete label foreground lies in $\sL_\tau\subseteq\sS$. Hence, a label-supported cell is retained even if its prediction value exceeds $\tau$, and that prediction value remains unchanged in the sparse filtration. Structures present in the label but appearing only late in the prediction filtration are therefore not discarded merely because of the sparsification threshold. More generally, for every inference threshold $\theta\leq\tau$, the prediction, label, and comparison complexes at $\theta$ are represented exactly. The resulting ordinary and image-persistence modules are used in the same Betti-Matching construction and with the same feature penalties as in the dense method \citep{stucki2023topologically}. If all three filtration functions are binary (e.g., for the computation of a metric between binarized images, studied in more detail in \Cref{sec:app_metric}), completion at level $1$ changes none of them, so the sparse and dense persistence data agree exactly.

\Cref{fig:barcode_example} shows an exemplary barcode with different cases (non-exhaustive) of how sparse and dense intervals might differ from each other in an image segmentation setting. Intervals (a), (b), and (e) show features with birth $<\tau$, where the exact birth is maintained. If death $<\tau$, it is kept exact (b). With death $>\tau$, it can be shifted (a) or kept exact (e) when the dense death is label-supported. Intervals with birth (and death) $>\tau$ can be shifted (case (c), e.g., when the shifted birth/death is label-supported) or entirely lost (d). A real-world example is depicted in \Cref{fig:gradient_similarity} (b).

\subsection{Implementation and design choices}
\label{sec:method_design}

\paragraph{Implicit representation of the omitted region.}
\label{sec:method_cones}
The implementation represents connected omitted regions by a virtual vertex and, where required, cone incidences over their interfaces with $\sS$. These incidences are generated implicitly rather than stored as an explicit complex. Union--find is used in the dimensions where persistence reduces to tracking connected components.

\paragraph{Sparsity and the choice of $\tau$.}
\label{sec:method_complexity}
Let $N$ be the number of cells in the full grid, $R$ the number of retained cells, and $I$ the size of the retained--omitted interface needed by the implicit representation. Constructing $\sS$ still requires a linear scan of the input, but the expensive persistence computations are governed primarily by $R+I$ rather than by $N$. Computational savings, therefore, require $R+I\ll N$, which is fulfilled in most segmentation settings (see \cref{fig:runtime}).
Notably, the bound $1-\tau$ requires no sparsity assumption, whereas the distribution of comparison-filtration values determines whether a large $\tau$ can simultaneously yield a small retained complex. In the segmentation setting, large confident background regions make this possible. In our experiments, we use $\tau=0.8$. Sparsification alters the persistence computation, while the feature penalties that define the Betti-Matching objective remain unchanged.

\section{Experiments}
Our empirical evaluation is structured into two parts to demonstrate the efficiency (\cref{sec:results_runtime}) and effectiveness (\cref{sec:results_performance}) of sparse cubical complexes in segmentation tasks. All of our analyses are oriented towards realistic imaging data and training paradigms.

\paragraph{Datasets.}
We choose six datasets (ATM~\citep{zhang2023multi}, NISB-B~\citep{rieger2024nisb}, BraTS-METS~\citep{maleki2025analysis}, MMWHS~\citep{Zhuang2016MSMMA}, FIVES~\citep{jin2022fives}, ACDC~\citep{bernard2018deep}) where topological correctness is important for downstream applications. These datasets cover various topological targets, such as vessel/airway connectivity (FIVES, ATM), loop closing (FIVES, ACDC), correctness of connected components (BraTS-METS), and closing of cavities (MMWHS, NISB). Although our method is focused on 3D applications, we include two 2D datasets (FIVES, ACDC) to enable comparison with additional baseline methods (see \cref{sec:results_performance}).

\subsection{Efficient PH computation with sparse cubical complexes}
\label{sec:results_runtime}

\begin{figure}[t]
    \centering
    \includegraphics[]{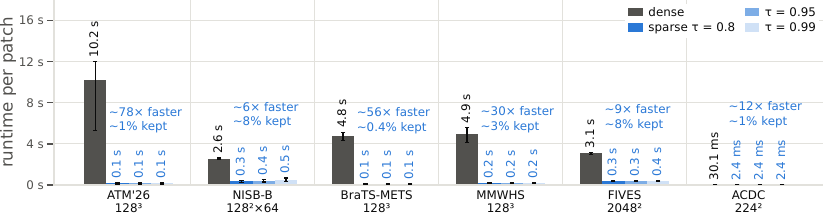}
    \caption{Barcode extraction time using dense and sparse cubical complexes on outputs of segmentation models.}
    \label{fig:runtime}
\end{figure}

In this section, we test whether the retained comparison subcomplex remains small on realistic segmentation outputs (fulfilling the complexity condition in \cref{sec:method_complexity}), and thus, whether our method yields runtime reductions in practice. We build cubical complexes from predictions during network training on the described datasets and extract barcodes via PH. We compare the extraction time between dense and sparse complexes with varying $\tau$ (\cref{fig:runtime}). First, we observe that all barcodes are extracted in less than half a second, making computation feasible in extremely time-critical settings, such as network training. Second, we find that PH computation is drastically faster on sparse cubical complexes than on full complexes. The speedup is especially pronounced on large patches with sparse foreground (e.g., $\times 78$ on ATM). The smallest measured speedup is $\times 6$ faster than dense on the NISB dataset, where patches are smaller, and foreground is more frequent. Third, we observe that the speedup scales roughly inversely with the kept fraction, which corroborates the computational description in \cref{sec:method_complexity}. The runtime is only marginally influenced by the choice of $\tau$, which is caused by a nearly constant kept fraction (see \Cref{sec:sup_tau_and_intensities}).

\subsection{SparseBM as an effective loss for image segmentation}
\label{sec:results_performance}

\begin{table*}[t]
\centering
\scriptsize
\setlength{\tabcolsep}{4.5pt}
\caption{Test-set results in 2D and 3D. Three seeds each, scored once on held-out data; mean$\pm$sd over seeds. Bold: best arm of the dataset. $^{*}$: sparse BM is significantly better than that arm on that metric (paired $t$-test over seeds, $p<0.05$). Train.\ time: wall clock of the whole training, as a multiple of Dice+CE. $^{\dagger}$: Runtime improved version of the authors' released code.}
\label{tab:main}
\begin{tabular}{llrrrrrr}
\toprule
dataset & loss & Dice$\uparrow$ & BM err.$\downarrow$ & clDice$\uparrow$ & VOI$\downarrow$ & NSD$\uparrow$ & Train.\ time$\downarrow$ \\
\midrule
\multirow{5}{*}{\begin{tabular}[c]{@{}l@{}}\textit{ATM'26}\\ {\tiny airway lumen}\\ {\tiny $128^3$}\end{tabular}} & Dice+CE & .9446{\tiny$\pm$.0032} & 102{\tiny$\pm$12}$^{*}$ & .900{\tiny$\pm$.002} & .0064{\tiny$\pm$.0003} & .955{\tiny$\pm$.004} & 1.00$\times$ \\
 & clDice & .9421{\tiny$\pm$.0048} & 84.8{\tiny$\pm$9.3}$^{*}$ & \textbf{.906}{\tiny$\pm$.006} & .0065{\tiny$\pm$.0004} & .953{\tiny$\pm$.004} & 1.29$\times$ \\
 & Skel.\ Recall & .9437{\tiny$\pm$.0022} & 96.4{\tiny$\pm$6.5}$^{*}$ & .896{\tiny$\pm$.003}$^{*}$ & .0065{\tiny$\pm$.0001}$^{*}$ & .953{\tiny$\pm$.001} & 1.03$\times$ \\
 & warping$^{\dagger}$ & .9413{\tiny$\pm$.0038} & 94.6{\tiny$\pm$12.3}$^{*}$ & .896{\tiny$\pm$.008} & .0066{\tiny$\pm$.0003} & .951{\tiny$\pm$.006} & 1.40$\times$ \\
 & \textbf{sparse BM (ours)} & \textbf{.9449}{\tiny$\pm$.0017} & \textbf{18.8}{\tiny$\pm$.5} & \textbf{.906}{\tiny$\pm$.006} & \textbf{.0061}{\tiny$\pm$.0001} & \textbf{.956}{\tiny$\pm$.004} & 1.15$\times$ \\
\midrule
\multirow{5}{*}{\begin{tabular}[c]{@{}l@{}}\textit{NISB-B}\\ {\tiny cell interfaces}\\ {\tiny $128^2{\times}64$}\end{tabular}} & Dice+CE & .7847{\tiny$\pm$.0001} & 367.3k{\tiny$\pm$3.3k}$^{*}$ & \textbf{.766}{\tiny$\pm$.000} & \textbf{4.521}{\tiny$\pm$.000} & .929{\tiny$\pm$.000} & 1.00$\times$ \\
 & clDice & .7708{\tiny$\pm$.0003}$^{*}$ & 208.7k{\tiny$\pm$706}$^{*}$ & .681{\tiny$\pm$.002}$^{*}$ & 4.563{\tiny$\pm$.001}$^{*}$ & \textbf{.930}{\tiny$\pm$.000} & 1.08$\times$ \\
 & Skel.\ Recall & .7841{\tiny$\pm$.0002} & 344.6k{\tiny$\pm$2.1k}$^{*}$ & .747{\tiny$\pm$.001}$^{*}$ & 4.531{\tiny$\pm$.001}$^{*}$ & .928{\tiny$\pm$.000} & 1.02$\times$ \\
 & warping$^{\dagger}$ & \textbf{.7851}{\tiny$\pm$.0001} & 373.1k{\tiny$\pm$2.1k}$^{*}$ & .763{\tiny$\pm$.000} & 4.522{\tiny$\pm$.000} & .929{\tiny$\pm$.000} & 1.09$\times$ \\
 & \textbf{sparse BM (ours)} & .7785{\tiny$\pm$.0001} & \textbf{80.0k}{\tiny$\pm$326} & .762{\tiny$\pm$.000} & 4.526{\tiny$\pm$.000} & .925{\tiny$\pm$.000} & 1.14$\times$ \\
\midrule
\multirow{5}{*}{\begin{tabular}[c]{@{}l@{}}\textit{BraTS-METS}\\ {\tiny NETC $\cup$ ET}\\ {\tiny $128^3$}\end{tabular}} & Dice+CE & .6854{\tiny$\pm$.0107} & 5.87{\tiny$\pm$.84}$^{*}$ & .673{\tiny$\pm$.013} & \textbf{.0069}{\tiny$\pm$.0003} & .766{\tiny$\pm$.007} & 1.00$\times$ \\
 & clDice & .6845{\tiny$\pm$.0126} & 5.33{\tiny$\pm$.42}$^{*}$ & .680{\tiny$\pm$.016} & .0070{\tiny$\pm$.0004} & .766{\tiny$\pm$.006} & 1.20$\times$ \\
 & Skel.\ Recall & .6885{\tiny$\pm$.0124} & 6.01{\tiny$\pm$.69}$^{*}$ & .657{\tiny$\pm$.057} & .0070{\tiny$\pm$.0005} & .759{\tiny$\pm$.023} & 1.01$\times$ \\
 & warping$^{\dagger}$ & .6785{\tiny$\pm$.0034} & 5.48{\tiny$\pm$.18}$^{*}$ & .663{\tiny$\pm$.023} & .0070{\tiny$\pm$.0002} & .762{\tiny$\pm$.017} & 1.29$\times$ \\
 & \textbf{sparse BM (ours)} & \textbf{.6889}{\tiny$\pm$.0112} & \textbf{4.16}{\tiny$\pm$.25} & \textbf{.701}{\tiny$\pm$.033} & .0074{\tiny$\pm$.0003} & \textbf{.770}{\tiny$\pm$.022} & 1.08$\times$ \\
\midrule
\multirow{5}{*}{\begin{tabular}[c]{@{}l@{}}\textit{MMWHS}\\ {\tiny LV myocardium}\\ {\tiny $128^3$}\end{tabular}} & Dice+CE & .8367{\tiny$\pm$.0107}$^{*}$ & 3.73{\tiny$\pm$1.68}$^{*}$ & .973{\tiny$\pm$.021}$^{*}$ & .0701{\tiny$\pm$.0025}$^{*}$ & .687{\tiny$\pm$.014}$^{*}$ & 1.00$\times$ \\
 & clDice & .8347{\tiny$\pm$.0113}$^{*}$ & 2.52{\tiny$\pm$.97} & .975{\tiny$\pm$.019} & .0710{\tiny$\pm$.0029}$^{*}$ & .680{\tiny$\pm$.016}$^{*}$ & 1.15$\times$ \\
 & Skel.\ Recall & \textbf{.8379}{\tiny$\pm$.0103} & 3.00{\tiny$\pm$1.15}$^{*}$ & \textbf{.976}{\tiny$\pm$.018} & .0699{\tiny$\pm$.0026} & .687{\tiny$\pm$.015}$^{*}$ & 1.22$\times$ \\
 & warping$^{\dagger}$ & .8313{\tiny$\pm$.0133} & 4.98{\tiny$\pm$1.56}$^{*}$ & .969{\tiny$\pm$.014}$^{*}$ & .0728{\tiny$\pm$.0047}$^{*}$ & .674{\tiny$\pm$.036} & 1.20$\times$ \\
 & \textbf{sparse BM (ours)} & .8377{\tiny$\pm$.0109} & \textbf{2.19}{\tiny$\pm$.94} & \textbf{.976}{\tiny$\pm$.020} & \textbf{.0697}{\tiny$\pm$.0028} & \textbf{.690}{\tiny$\pm$.017} & 1.05$\times$ \\
\midrule
\midrule
\multirow{7}{*}{\begin{tabular}[c]{@{}l@{}}\textit{FIVES (2D)}\\ {\tiny retinal vessels}\\ {\tiny $2048^2$}\end{tabular}} & Dice+CE & .9145{\tiny$\pm$.0001} & 64.7{\tiny$\pm$.2}$^{*}$ & .912{\tiny$\pm$.000} & \textbf{.154}{\tiny$\pm$.001} & \textbf{.811}{\tiny$\pm$.000} & 1.00$\times$ \\
 & clDice & .9140{\tiny$\pm$.0001} & 59.4{\tiny$\pm$.3}$^{*}$ & \textbf{.913}{\tiny$\pm$.000} & .155{\tiny$\pm$.000} & .809{\tiny$\pm$.000} & 1.09$\times$ \\
 & Skel.\ Recall & .9113{\tiny$\pm$.0003} & 62.3{\tiny$\pm$.6}$^{*}$ & .909{\tiny$\pm$.001} & .161{\tiny$\pm$.000}$^{*}$ & .799{\tiny$\pm$.001} & 1.04$\times$ \\
 & warping$^{\dagger}$ & \textbf{.9146}{\tiny$\pm$.0001} & 64.4{\tiny$\pm$.2}$^{*}$ & \textbf{.913}{\tiny$\pm$.000} & .155{\tiny$\pm$.000} & \textbf{.811}{\tiny$\pm$.000} & 2.11$\times$ \\
 & DMT$^{\dagger}$ & .9108{\tiny$\pm$.0005} & 55.2{\tiny$\pm$.3}$^{*}$ & .912{\tiny$\pm$.001} & .158{\tiny$\pm$.001}$^{*}$ & .796{\tiny$\pm$.001}$^{*}$ & 20.12$\times$ \\
 & Topograph & .9105{\tiny$\pm$.0001} & \textbf{40.1}{\tiny$\pm$1.0} & .912{\tiny$\pm$.000} & \textbf{.154}{\tiny$\pm$.001} & .801{\tiny$\pm$.001} & 1.37$\times$ \\
 & \textbf{sparse BM (ours)} & .9106{\tiny$\pm$.0007} & 40.3{\tiny$\pm$.7} & .911{\tiny$\pm$.001} & \textbf{.154}{\tiny$\pm$.001} & .798{\tiny$\pm$.001} & 1.40$\times$ \\
\midrule
\multirow{7}{*}{\begin{tabular}[c]{@{}l@{}}\textit{ACDC (2D)}\\ {\tiny LV myocardium}\\ {\tiny $224^2$}\end{tabular}} & Dice+CE & .8928{\tiny$\pm$.0036} & .241{\tiny$\pm$.016}$^{*}$ & .974{\tiny$\pm$.003}$^{*}$ & .0364{\tiny$\pm$.0000} & .907{\tiny$\pm$.002} & 1.00$\times$ \\
 & clDice & .8925{\tiny$\pm$.0028} & .116{\tiny$\pm$.019} & .979{\tiny$\pm$.003} & .0357{\tiny$\pm$.0003} & \textbf{.914}{\tiny$\pm$.002} & 1.06$\times$ \\
 & Skel.\ Recall & \textbf{.8979}{\tiny$\pm$.0010} & .218{\tiny$\pm$.022}$^{*}$ & \textbf{.980}{\tiny$\pm$.003} & .0362{\tiny$\pm$.0002} & .912{\tiny$\pm$.003} & 1.00$\times$ \\
 & warping$^{\dagger}$ & .8949{\tiny$\pm$.0012} & .244{\tiny$\pm$.021}$^{*}$ & .975{\tiny$\pm$.002}$^{*}$ & .0359{\tiny$\pm$.0001} & .911{\tiny$\pm$.002} & 1.35$\times$ \\
 & DMT$^{\dagger}$ & .8969{\tiny$\pm$.0028} & .171{\tiny$\pm$.031}$^{*}$ & .976{\tiny$\pm$.003} & \textbf{.0355}{\tiny$\pm$.0003} & \textbf{.914}{\tiny$\pm$.004} & 8.02$\times$ \\
 & Topograph & .8874{\tiny$\pm$.0027}$^{*}$ & .133{\tiny$\pm$.033} & .977{\tiny$\pm$.003} & .0365{\tiny$\pm$.0007} & .909{\tiny$\pm$.008} & 1.10$\times$ \\
 & \textbf{sparse BM (ours)} & .8937{\tiny$\pm$.0010} & \textbf{.109}{\tiny$\pm$.008} & \textbf{.980}{\tiny$\pm$.003} & .0364{\tiny$\pm$.0008} & .912{\tiny$\pm$.007} & 1.68$\times$ \\
\bottomrule
\end{tabular}
\end{table*}

In this section, we evaluate our proposed approach as a segmentation loss, sparseBM, and showcase its efficiency and effectiveness in reducing topological errors in state-of-the-art segmentation pipelines. We first briefly describe our experimental setup before presenting the core results. Lastly, we compare sparseBM to its dense counterpart (Betti Matching) mechanistically via its gradients and in a reduced experimental setting where dense PH computation is feasible for network training.

\subsubsection{Training setup.}
\label{sec:training_setup}
One of this paper's main objectives is to make PH-based loss functions applicable to real-world image segmentation pipelines. We want to move away from constructed, small-scale scenarios in which the proposed methods work and show improvements, towards the staple of image segmentation, specifically in 3D. Therefore, we follow the training principles of nnUnet, which has proven to be one of the most robust and versatile image segmentation \emph{recipes} for 3D images \citep{isensee2021nnu,isensee2024nnu}. At its core, nnUnet achieves its high performance by making appropriate choices for data preprocessing and model architecture, applying heavy augmentations, maximizing patch size, and training with a polynomial decaying learning rate.

\paragraph{Baselines.}
In our 3D experiments, we compare our method to the ComboLoss (Dice+CE) baseline, clDice \citep{shit2021cldice}, the skeleton recall loss \citep{kirchhoff2024skeleton}, and homotopy warping \citep{hu2022structure}. Other frequently used approaches are either not applicable to 3D (e.g., Topograph \citep{lux2025topograph}) or have runtime constraints making them infeasible in our setting (e.g., Betti Matching \citep{stucki2023topologically,stucki2024efficient}, or DMT \citep{hu2021topology}). To enable further comparisons, we add two 2D datasets (comparing to Topograph and DMT) and training with reduced patch size (\Cref{tab:dense}, comparing to DMT and Betti Matching) to our experiments. Furthermore, we optimize the implementation of homotopy warping and DMT to reduce runtime.

\paragraph{Metrics.}
The empirical evaluation should answer one main question: can our method reliably reduce topological errors while maintaining pixel- and region-wise accuracy? For this, we consider two main metrics: Dice score and Betti-matching (BM) error using 6-connectivity of the binarized images. Following \citet{berger2025pitfalls}, BM error is the most suitable topological metric as it measures \emph{pure} topology, i.e., it is not diluted by other factors, such as skeletonization, region sizes, or pixel-wise accuracy. However, to make our benchmark comparable to prior work, we report clDice, Variation of Information (VOI) \citep{meilua2003comparing}, and Normalized Surface Distance (NSD) \citep{nikolov2018deep}. While these metrics are appropriate in specific domains (e.g., tubular structures or boundary segmentation), they lack general applicability and should therefore be considered only as additional information \citep{berger2025pitfalls}. All metrics are computed on the entire test volumes via sliding window inference \citep{isensee2021nnu}.

\subsubsection{Results}

Sparse cubical complexes, when used as the foundation for computing Betti Matching (i.e., \textbf{sparseBM}), substantially reduce topological errors compared to the ComboLoss across all six datasets while keeping Dice within the prespecified one-percentage-point admissibility criterion used for model selection. We observe up to a 5-fold reduction in BM-error with minimal runtime overhead between $5$ and $15\%$ on 3D data. Similarly, sparseBM yields significant topological improvements over clDice and skeleton recall on ATM'26, NISB-B, and BraTS-METS. On MMWHS, clDice achieves comparable results but incurs a $15\%$ runtime overhead, compared to sparseBM's $5\%$. We do not observe meaningful changes in VOI, NSD, or the clDice metric across methods on any dataset. This is caused by confounding factors that influence these metrics, such as foreground and region sizes, as described in \citep{berger2025pitfalls}.

In our 2D experiments, with an extended set of baselines covering DMT and Topograph, sparseBM achieves a significant reduction in topological errors compared to all baselines except Topograph (on both datasets) and clDice (with an insignificant improvement only on ACDC). sparseBM's runtime overhead on 2D data is higher because of a larger foreground fraction. The time spent on network training itself (forward and backward passes on the GPU) is smaller, so the loss calculation takes up a larger fraction of the total training time.

\paragraph{Functional similarity to Betti Matching}
\begin{figure}[t]
    \centering
    \includegraphics[width=\linewidth]{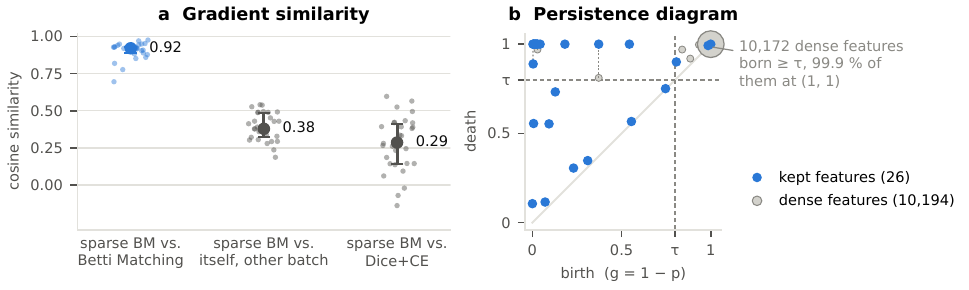}
    \caption{SparseBM closely reproduces the optimization signal of dense Betti Matching. (a) shows the cosine similarity between their update steps, together with baseline comparisons to ComboLoss and to a different batch with the same loss. (b) shows the corresponding difference in persistence diagrams. Most affected features have very small persistence and lie in the background. Both plots are created from real samples during training on the ATM'26 dataset.}
    \label{fig:gradient_similarity}
\end{figure}

\begin{table*}[t]
\centering
\scriptsize
\setlength{\tabcolsep}{4.5pt}
\caption{Comparison to PH-based losses in a training setting with reduced patch size on the ATM'26 dataset. Same annotation as in \Cref{tab:main}.}
\label{tab:dense}
\begin{tabular}{llrrrrrr}
\toprule
patch & loss & Dice$\uparrow$ & BM err.$\downarrow$ & clDice$\uparrow$ & VOI$\downarrow$ & NSD$\uparrow$ & Train.\ time$\downarrow$ \\
\midrule
\multirow{2}{*}{$128^3$} & Dice+CE & .9446{\tiny$\pm$.0032} & 102{\tiny$\pm$12}$^{*}$ & .900{\tiny$\pm$.002} & .0064{\tiny$\pm$.0003} & .955{\tiny$\pm$.004} & 1.00$\times$ \\
 & \textbf{sparse BM (ours)} & \textbf{.9449}{\tiny$\pm$.0017} & \textbf{18.8}{\tiny$\pm$.5} & \textbf{.906}{\tiny$\pm$.006} & \textbf{.0061}{\tiny$\pm$.0001} & \textbf{.956}{\tiny$\pm$.004} & 1.15$\times$ \\
\midrule
\multirow{4}{*}{$64^3$} & Dice+CE & .9299{\tiny$\pm$.0124} & 129{\tiny$\pm$20}$^{*}$ & .878{\tiny$\pm$.014}$^{*}$ & .0075{\tiny$\pm$.0011} & .939{\tiny$\pm$.013} & 1.00$\times$ \\
 & DMT & .9407{\tiny$\pm$.0013} & 66.5{\tiny$\pm$1.8}$^{*}$ & .906{\tiny$\pm$.004} & .0065{\tiny$\pm$.0001} & \textbf{.957}{\tiny$\pm$.002} & 5.38$\times$ \\
 & dense BM & .9358{\tiny$\pm$.0062} & 25.1{\tiny$\pm$1.8}$^{*}$ & .904{\tiny$\pm$.002}$^{*}$ & .0069{\tiny$\pm$.0005} & .949{\tiny$\pm$.004} & 1.88$\times$ \\
 & \textbf{sparse BM (ours)} & \textbf{.9412}{\tiny$\pm$.0023} & \textbf{20.2}{\tiny$\pm$2.1} & \textbf{.909}{\tiny$\pm$.000} & \textbf{.0063}{\tiny$\pm$.0001} & .954{\tiny$\pm$.001} & 1.19$\times$ \\
\bottomrule
\end{tabular}
\end{table*}

Given the large runtime difference between sparseBM and dense PH-based loss functions, we cannot make a direct comparison in state-of-the-art training pipelines. Therefore, we run two alternative experiments.

First, we use network predictions from training on the ATM'26 dataset and compute the gradients of sparseBM and Betti Matching. We compute the mean cosine similarity between the respective update steps and find high similarity (\cref{fig:gradient_similarity}.a). For comparison, we compute the mean cosine similarity between the update steps of two subsequent batches with the same loss and of sparseBM and the ComboLoss. This high similarity is consistent with the theoretical localization of the sparse--dense discrepancy to the terminal filtration band above $\tau$, whose width is $1-\tau$, and with the empirical observation in \cref{fig:gradient_similarity}.b that most affected features in this example have almost zero persistence and lie in the background. Individually, such features make only a small persistence-based contribution, while processing their cells accounts for much of the dense runtime described in \cref{sec:results_runtime}.

Furthermore, we repeat the training experiment with a reduced patch size, enabling dense PH computation on the ATM'26 dataset (\cref{tab:dense}). First, we note that smaller patch sizes reduce topological and pixel-wise accuracy. Additionally, we find that both Betti Matching and sparseBM drastically ($\times 5$ and $\times 6$, respectively) reduce topological errors compared to the ComboLoss baseline without compromising on the Dice score. At the same time, sparseBM achieves a significant reduction in BM error compared to Betti Matching. We hypothesize that omitting this terminal-band information may even be beneficial for segmentation networks. PH-based losses generally suppress all unmatched features, i.e., they enforce a smooth intensity landscape, even in confident foreground and background regions. This characteristic has no effect on binarized segmentation maps and therefore, might \emph{distract} during the training process. Even in this reduced setting, sparseBM is substantially faster than Betti Matching with a total runtime overhead of $19\%$ compared with $88\%$. Notably, the runtime overhead of sparseBM remains roughly constant across the two patch sizes considered here, illustrating its reduced sensitivity to patch size in this sparse regime.

\section{Conclusion}
We propose sparse cubical complexes that omit high-valued regions while retaining the filtration values on a common comparison-based subcomplex. For topological losses in image segmentation, this yields sparseBM, a sparse variant of Betti Matching. In our experiments, sparseBM reduces loss-computation time by up to $\times100$ and closely reproduces the optimization signal of its dense counterpart, as indicated by the similarity of their loss gradients. Across six datasets, including four 3D large-patch datasets for which dense PH computation during training is practically infeasible, we observe substantial improvements in topological segmentation accuracy.

\paragraph{Limitations and future work.}
The computational advantage of the sparse complex depends on the retained comparison subcomplex and its interface remaining small. In the segmentation tasks considered here, this is typically enabled by large confident background regions. Speed-ups will be smaller when a large fraction of the comparison filtration lies below the sparsification threshold.
More generally, topology-preserving segmentation losses have the inherent limitation of providing only train-time guarantees that do not generalize during inference. The application of sparse cubical filtrations to other imaging tasks, in which runtime is critical, such as classification, generation, or reconstruction, warrants future work. In \Cref{sec:app_other}, we present our approach's efficiency for use as a metric and the first results for its use as a post-processing method, scoring 3rd in the official ATM'26 challenge without any other additional improvements.

\subsection*{Reproducibility statement}
The implementation of sparse cubical complexes and of sparseBM (C++ persistence and
matching with Python bindings, and the PyTorch loss) is available at
\url{https://github.com/AlexanderHBerger/sparse-cubical-filtration}. The theoretical statements in
\Cref{sec:method} follow from the induced matching theorem \citep{bauer2014induced}. All six
datasets are publicly available; \Cref{sec:app_preprocessing,sec:app_datasets} and \Cref{tab:datasets} give the
preprocessing, exclusions and train/validation/test splits. The training recipe
(\Cref{sec:training_setup}), the model-selection protocol (loss-weight search on the validation
fold, one-point Dice admissibility criterion, three seeds, a single scoring of the
held-out test set, paired $t$-tests) and the ablations on $\lambda$ and $\tau$ are
described in \Cref{sec:results_performance} and \Crefrange{sec:app_weight_ablation}{sec:app_experimental_design}. Baselines use the authors' released
code; homotopy warping and DMT use runtime-optimized versions of it (marked $\dagger$).

\bibliography{references}

@inproceedings{stucki2023topologically,
  title={Topologically faithful image segmentation via induced matching of persistence barcodes},
  author={Stucki, Nico and Paetzold, Johannes C and Shit, Suprosanna and Menze, Bjoern and Bauer, Ulrich},
  booktitle={International Conference on Machine Learning},
  pages={32698--32727},
  year={2023},
  organization={PMLR}
}

@article{edelsbrunner2008persistent,
  title={Persistent homology-a survey},
  author={Edelsbrunner, Herbert and Harer, John and others},
  journal={Contemporary mathematics},
  volume={453},
  number={26},
  pages={257--282},
  year={2008},
  publisher={Providence, RI: American Mathematical Society}
}

@inproceedings{heiss2017streaming,
  title={Streaming algorithm for Euler characteristic curves of multidimensional images},
  author={Heiss, Teresa and Wagner, Hubert},
  booktitle={International Conference on Computer Analysis of Images and Patterns},
  pages={397--409},
  year={2017},
  organization={Springer}
}

@article{stucki2024efficient,
  title={Efficient betti matching enables topology-aware 3d segmentation via persistent homology},
  author={Stucki, Nico and B{\"u}rgin, Vincent and Paetzold, Johannes C and Bauer, Ulrich},
  journal={arXiv preprint arXiv:2407.04683},
  year={2024}
}

@article{hu2019topology,
  title={Topology-preserving deep image segmentation},
  author={Hu, Xiaoling and Li, Fuxin and Samaras, Dimitris and Chen, Chao},
  journal={Advances in neural information processing systems},
  volume={32},
  year={2019}
}

@article{clough2020topological,
  title={A topological loss function for deep-learning based image segmentation using persistent homology},
  author={Clough, James R and Byrne, Nicholas and Oksuz, Ilkay and Zimmer, Veronika A and Schnabel, Julia A and King, Andrew P},
  journal={IEEE transactions on pattern analysis and machine intelligence},
  volume={44},
  number={12},
  pages={8766--8778},
  year={2020},
  publisher={IEEE}
}

@inproceedings{clough2019explicit,
  title={Explicit topological priors for deep-learning based image segmentation using persistent homology},
  author={Clough, James R and Oksuz, Ilkay and Byrne, Nicholas and Schnabel, Julia A and King, Andrew P},
  booktitle={International Conference on Information Processing in Medical Imaging},
  pages={16--28},
  year={2019},
  organization={Springer}
}

@inproceedings{berger2024topologically,
  title={Topologically faithful multi-class segmentation in medical images},
  author={Berger, Alexander H and Lux, Laurin and Stucki, Nico and B{\"u}rgin, Vincent and Shit, Suprosanna and Banaszak, Anna and Rueckert, Daniel and Bauer, Ulrich and Paetzold, Johannes C},
  booktitle={International Conference on Medical Image Computing and Computer-Assisted Intervention},
  pages={721--731},
  year={2024},
  organization={Springer}
}

@inproceedings{chen2011persistent,
  title={Persistent homology computation with a twist},
  author={Chen, Chao and Kerber, Michael},
  booktitle={Proceedings 27th European workshop on computational geometry},
  volume={11},
  pages={197--200},
  year={2011}
}

@article{bauer2021ripser,
  title={Ripser: efficient computation of Vietoris--Rips persistence barcodes},
  author={Bauer, Ulrich},
  journal={Journal of Applied and Computational Topology},
  volume={5},
  number={3},
  pages={391--423},
  year={2021},
  publisher={Springer}
}

@inproceedings{carriere2021optimizing,
  title={Optimizing persistent homology based functions},
  author={Carriere, Mathieu and Chazal, Fr{\'e}d{\'e}ric and Glisse, Marc and Ike, Yuichi and Kannan, Hariprasad and Umeda, Yuhei},
  booktitle={International conference on machine learning},
  pages={1294--1303},
  year={2021},
  organization={PMLR}
}

@inproceedings{bauer2014induced,
  title={Induced matchings of barcodes and the algebraic stability of persistence},
  author={Bauer, Ulrich and Lesnick, Michael},
  booktitle={Proceedings of the thirtieth annual symposium on Computational geometry},
  pages={355--364},
  year={2014}
}

@article{kaji2020cubical,
  title={Cubical ripser: Software for computing persistent homology of image and volume data},
  author={Kaji, Shizuo and Sudo, Takeki and Ahara, Kazushi},
  journal={arXiv preprint arXiv:2005.12692},
  year={2020}
}

@inproceedings{hu2021topology,
  title={Topology-aware segmentation using discrete morse theory},
  author={Hu, Xiaoling and Wang, Yusu and Fuxin, Li and Samaras, Dimitris and Chen, Chao},
  booktitle={International Conference on Learning Representations},
  volume={2021},
  pages={2101},
  year={2021}
}

@inproceedings{shit2021cldice,
  title={clDice-a novel topology-preserving loss function for tubular structure segmentation},
  author={Shit, Suprosanna and Paetzold, Johannes C and Sekuboyina, Anjany and Ezhov, Ivan and Unger, Alexander and Zhylka, Andrey and Pluim, Josien PW and Bauer, Ulrich and Menze, Bjoern H},
  booktitle={2021 IEEE/CVF Conference on Computer Vision and Pattern Recognition (CVPR)},
  pages={16555--16564},
  year={2021},
  organization={IEEE}
}

@inproceedings{kirchhoff2024skeleton,
  title={Skeleton recall loss for connectivity conserving and resource efficient segmentation of thin tubular structures},
  author={Kirchhoff, Yannick and Rokuss, Maximilian R and Roy, Saikat and Kovacs, Balint and Ulrich, Constantin and Wald, Tassilo and Zenk, Maximilian and Vollmuth, Philipp and Kleesiek, Jens and Isensee, Fabian and others},
  booktitle={European Conference on Computer Vision},
  pages={218--234},
  year={2024},
  organization={Springer}
}

@inproceedings{lux2025topograph,
  title={Topograph: An efficient graph-based framework for strictly topology preserving image segmentation},
  author={Lux, Laurin and Berger, Alexander H and Weers, Alexander and Stucki, Nico and Rueckert, Daniel and Bauer, Ulrich and Paetzold, Johannes},
  booktitle={International Conference on Learning Representations},
  volume={2025},
  pages={80207--80231},
  year={2025}
}

@inproceedings{berger2025pitfalls,
  title={Pitfalls of topology-aware image segmentation},
  author={Berger, Alexander H and Lux, Laurin and Weers, Alexander and Menten, Martin J and Rueckert, Daniel and Paetzold, Johannes C},
  booktitle={International Conference on Information Processing in Medical Imaging},
  pages={297--312},
  year={2025},
  organization={Springer}
}

@inproceedings{menten2023skeletonization,
  title={A skeletonization algorithm for gradient-based optimization},
  author={Menten, Martin J and Paetzold, Johannes C and Zimmer, Veronika A and Shit, Suprosanna and Ezhov, Ivan and Holland, Robbie and Probst, Monika and Schnabel, Julia A and Rueckert, Daniel},
  booktitle={2023 IEEE/CVF International Conference on Computer Vision (ICCV)},
  pages={21337--21346},
  year={2023},
  organization={IEEE}
}

@inproceedings{isensee2024nnu,
  title={nnu-net revisited: A call for rigorous validation in 3d medical image segmentation},
  author={Isensee, Fabian and Wald, Tassilo and Ulrich, Constantin and Baumgartner, Michael and Roy, Saikat and Maier-Hein, Klaus and Jaeger, Paul F},
  booktitle={International conference on medical image computing and computer-assisted intervention},
  pages={488--498},
  year={2024},
  organization={Springer}
}

@article{isensee2021nnu,
  title={nnU-Net: a self-configuring method for deep learning-based biomedical image segmentation},
  author={Isensee, Fabian and Jaeger, Paul F and Kohl, Simon AA and Petersen, Jens and Maier-Hein, Klaus H},
  journal={Nature methods},
  volume={18},
  number={2},
  pages={203--211},
  year={2021},
  publisher={Nature Publishing Group US New York}
}

@inproceedings{moor2020topological,
  title={Topological autoencoders},
  author={Moor, Michael and Horn, Max and Rieck, Bastian and Borgwardt, Karsten},
  booktitle={International conference on machine learning},
  pages={7045--7054},
  year={2020},
  organization={PMLR}
}

@inproceedings{gupta2025topodiffusionnet,
  title={Topodiffusionnet: A topology-aware diffusion model},
  author={Gupta, Saumya and Samaras, Dimitris and Chen, Chao},
  booktitle={International Conference on Learning Representations},
  volume={2025},
  pages={31699--31713},
  year={2025}
}

@inproceedings{qi2023dynamic,
  title={Dynamic snake convolution based on topological geometric constraints for tubular structure segmentation},
  author={Qi, Yaolei and He, Yuting and Qi, Xiaoming and Zhang, Yuan and Yang, Guanyu},
  booktitle={2023 IEEE/CVF International Conference on Computer Vision (ICCV)},
  pages={6047--6056},
  year={2023},
  organization={IEEE}
}

@article{byrne2022persistent,
  title={A persistent homology-based topological loss for CNN-based multiclass segmentation of CMR},
  author={Byrne, Nick and Clough, James R and Valverde, Israel and Montana, Giovanni and King, Andrew P},
  journal={IEEE transactions on medical imaging},
  volume={42},
  number={1},
  pages={3--14},
  year={2022},
  publisher={IEEE}
}

@article{qaiser2019fast,
  title={Fast and accurate tumor segmentation of histology images using persistent homology and deep convolutional features},
  author={Qaiser, Talha and Tsang, Yee-Wah and Taniyama, Daiki and Sakamoto, Naoya and Nakane, Kazuaki and Epstein, David and Rajpoot, Nasir},
  journal={Medical image analysis},
  volume={55},
  pages={1--14},
  year={2019},
  publisher={Elsevier}
}

@inproceedings{xu2025topocellgen,
  title={Topocellgen: Generating histopathology cell topology with a diffusion model},
  author={Xu, Meilong and Gupta, Saumya and Hu, Xiaoling and Li, Chen and Abousamra, Shahira and Samaras, Dimitris and Prasanna, Prateek and Chen, Chao},
  booktitle={2025 IEEE/CVF Conference on Computer Vision and Pattern Recognition (CVPR)},
  pages={20979--20989},
  year={2025},
  organization={IEEE}
}

@article{funke2018large,
  title={Large scale image segmentation with structured loss based deep learning for connectome reconstruction},
  author={Funke, Jan and Tschopp, Fabian and Grisaitis, William and Sheridan, Arlo and Singh, Chandan and Saalfeld, Stephan and Turaga, Srinivas C},
  journal={IEEE transactions on pattern analysis and machine intelligence},
  volume={41},
  number={7},
  pages={1669--1680},
  year={2018},
  publisher={IEEE}
}

@article{alastruey2007modelling,
  title={Modelling the circle of Willis to assess the effects of anatomical variations and occlusions on cerebral flows},
  author={Alastruey, JPKH and Parker, Kim H and Peiro, Joaquim and Byrd, Shawn M and Sherwin, Spencer J},
  journal={Journal of biomechanics},
  volume={40},
  number={8},
  pages={1794--1805},
  year={2007},
  publisher={Elsevier}
}

@article{hu2022structure,
  title={Structure-aware image segmentation with homotopy warping},
  author={Hu, Xiaoling},
  journal={Advances in Neural Information Processing Systems},
  volume={35},
  pages={24046--24059},
  year={2022}
}

@inproceedings{meilua2003comparing,
  title={Comparing clusterings by the variation of information},
  author={Meil{\u{a}}, Marina},
  booktitle={Learning Theory and Kernel Machines: 16th Annual Conference on Learning Theory and 7th Kernel Workshop, COLT/Kernel 2003, Washington, DC, USA, August 24-27, 2003. Proceedings},
  pages={173--187},
  year={2003},
  organization={Springer}
}

@article{nikolov2018deep,
  title={Deep learning to achieve clinically applicable segmentation of head and neck anatomy for radiotherapy},
  author={Nikolov, Stanislav and Blackwell, Sam and Zverovitch, Alexei and Mendes, Ruheena and Livne, Michelle and De Fauw, Jeffrey and Patel, Yojan and Meyer, Clemens and Askham, Harry and Romera-Paredes, Bernardino and others},
  journal={arXiv preprint arXiv:1809.04430},
  year={2018}
}

@inproceedings{wang2020topogan,
  title={Topogan: A topology-aware generative adversarial network},
  author={Wang, Fan and Liu, Huidong and Samaras, Dimitris and Chen, Chao},
  booktitle={European Conference on Computer Vision},
  pages={118--136},
  year={2020},
  organization={Springer}
}

@article{hofer2017deep,
  title={Deep learning with topological signatures},
  author={Hofer, Christoph and Kwitt, Roland and Niethammer, Marc and Uhl, Andreas},
  journal={Advances in neural information processing systems},
  volume={30},
  year={2017}
}

@article{breton2026fast,
  title={Fast Cubical Persistent Homology on 2D and 3D Images via Union-Find, Pruning, and Lookup Tables},
  author={Breton, Titouan Le and Szustakowski, Karol and Piraud, Marie},
  journal={arXiv preprint arXiv:2606.04801},
  year={2026}
}

@article{zhang2023multi,
  title={Multi-site, multi-domain airway tree modeling},
  author={Zhang, Minghui and Wu, Yangqian and Zhang, Hanxiao and Qin, Yulei and Zheng, Hao and Tang, Wen and Arnold, Corey and Pei, Chenhao and Yu, Pengxin and Nan, Yang and others},
  journal={Medical image analysis},
  volume={90},
  pages={102957},
  year={2023},
  publisher={Elsevier}
}

@misc{rieger2024nisb,
  doi = {10.17617/1.R2MM-1H33},
  url = {https://structuralneurobiologylab.github.io/nisb/},
  author = {Rieger, Franz and Lăcătușu, Ana-Maria and Urbanová, Zuzana and Mancu, Andrei and Ahmad, Hashir and Bucella, Martin and Kornfeld, Joergen},
  title = {NISB: Neuron Instance Segmentation Benchmark},
  publisher = {Max Planck Institute for Biological Intelligence},
  year = {2024}
}

@article{maleki2025analysis,
  title={Analysis of the MICCAI brain tumor segmentation--metastases (BraTS-METS) 2025 lighthouse challenge: brain metastasis segmentation on pre-and post-treatment MRI},
  author={Maleki, Nazanin and Amiruddin, Raisa and Moawad, Ahmed W and Yordanov, Nikolay and Gkampenis, Athanasios and Fehringer, Pascal and Umeh, Fabian and Chukwurah, Crystal and Memon, Fatima and Petrovic, Bojan and others},
  journal={arXiv preprint arXiv:2504.12527},
  year={2025}
}

@article{Zhuang2016MSMMA,
  Author = {Zhuang, Xiahai and Shen, Juan},
  Title = {Multi-scale patch and multi-modality atlases for whole heart
     segmentation of MRI},
  Journal = {Medical Image Analysis},
  Year = {2016},
  Volume = {31},
  Pages = {77-87},
}

@article{jin2022fives,
  title={Fives: A fundus image dataset for artificial intelligence based vessel segmentation},
  author={Jin, Kai and Huang, Xingru and Zhou, Jingxing and Li, Yunxiang and Yan, Yan and Sun, Yibao and Zhang, Qianni and Wang, Yaqi and Ye, Juan},
  journal={Scientific data},
  volume={9},
  number={1},
  pages={475},
  year={2022},
  publisher={Nature Publishing Group UK London}
}

@article{bernard2018deep,
  title={Deep learning techniques for automatic MRI cardiac multi-structures segmentation and diagnosis: is the problem solved?},
  author={Bernard, Olivier and Lalande, Alain and Zotti, Clement and Cervenansky, Frederick and Yang, Xin and Heng, Pheng-Ann and Cetin, Irem and Lekadir, Karim and Camara, Oscar and Ballester, Miguel Angel Gonzalez and others},
  journal={IEEE transactions on medical imaging},
  volume={37},
  number={11},
  pages={2514--2525},
  year={2018},
  publisher={ieee}
}

@article{lawson2019persistent,
  title={Persistent homology for the quantitative evaluation of architectural features in prostate cancer histology},
  author={Lawson, Peter and Sholl, Andrew B and Brown, J Quincy and Fasy, Brittany Terese and Wenk, Carola},
  journal={Scientific reports},
  volume={9},
  number={1},
  pages={1139},
  year={2019},
  publisher={Nature Publishing Group UK London}
}

@article{li2025topology,
  title={Topology optimization in medical image segmentation with fast $\chi$ euler characteristic},
  author={Li, Liu and Ma, Qiang and Ouyang, Cheng and Paetzold, Johannes C and Rueckert, Daniel and Kainz, Bernhard},
  journal={IEEE Transactions on Medical Imaging},
  volume={44},
  number={12},
  pages={5221--5232},
  year={2025},
  publisher={IEEE}
}

@article{oner2023persistent,
  title={Persistent homology with improved locality information for more effective delineation},
  author={Oner, Doruk and Garin, Ad{\'e}lie and Kozi{\'n}ski, Mateusz and Hess, Kathryn and Fua, Pascal},
  journal={IEEE Transactions on Pattern Analysis and Machine Intelligence},
  volume={45},
  number={8},
  pages={10588--10595},
  year={2023},
  publisher={IEEE}
}

@inproceedings{wen2025topology,
  title={Topology-preserving image segmentation with spatial-aware persistent feature matching},
  author={Wen, Bo and Zhang, Haochen and Bartsch, Dirk-Uwe G and Freeman, William and Nguyen, Truong and An, Cheolhong},
  booktitle={Proceedings of the IEEE/CVF International Conference on Computer Vision},
  pages={5821--5830},
  year={2025}
}

@article{valverde2026towards,
  title={Towards High-Quality Image Segmentation: Improving Topology Accuracy by Penalizing Neighbor Pixels},
  author={Valverde, Juan Miguel and Papadopoulos, Dim P and Larsen, Rasmus and Dahl, Anders Bjorholm},
  journal={arXiv preprint arXiv:2603.18671},
  year={2026}
}

@article{ma2026topology,
  title={Topology-Preserving retinal vascular segmentation via sparse persistent homology and MoE convolution},
  author={Ma, Benteng and Li, Xiaomeng and Pu, Bin and Cheng, Kwang-Ting},
  journal={Pattern Recognition},
  pages={113612},
  year={2026},
  publisher={Elsevier}
}

@article{xu2026match,
  title={Match: Multi-faceted adaptive topo-consistency for semi-supervised histopathology segmentation},
  author={Xu, Meilong and Hu, Xiaoling and Abousamra, Shahira and Li, Chen and Chen, Chao},
  journal={Advances in Neural Information Processing Systems},
  volume={38},
  pages={105646--105672},
  year={2026}
}

@article{nigmetov2024big,
  title={Topological optimization with big steps},
  author={Nigmetov, Arnur and Morozov, Dmitriy},
  journal={Discrete \& computational geometry},
  volume={72},
  number={1},
  pages={310--344},
  year={2024},
  publisher={Springer}
}

@inproceedings{cohen2009persistent,
  title={Persistent homology for kernels, images, and cokernels},
  author={Cohen-Steiner, David and Edelsbrunner, Herbert and Harer, John and Morozov, Dmitriy},
  booktitle={Proceedings of the twentieth annual ACM-SIAM symposium on Discrete algorithms},
  pages={1011--1020},
  year={2009},
  organization={SIAM}
}
\bibliographystyle{iclr2027_conference}

\appendix
\section{Appendix}

\subsection{Ablation on weight parameter}
\label{sec:app_weight_ablation}

\begin{figure}[h]
    \centering
    \includegraphics[width=0.5\linewidth]{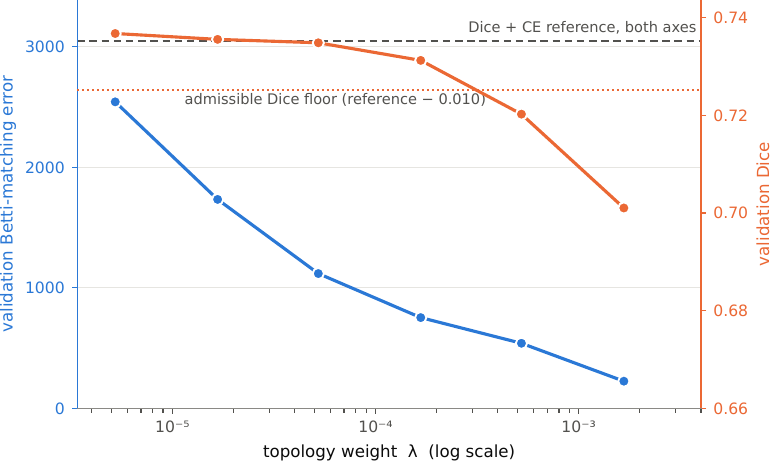}
    \caption{Ablation on the weight parameter on the NISB-B dataset (validation set, individual patches). Betti Matching error (left axis, blue) and Dice (right axis, orange) against the topology weight. The dashed line is the Dice+CE baseline, the dotted line the admissibility floor (reference $-\,0.010$) for model selection. The selected weight is the last point above the floor.
    }
    \label{fig:weight_ablation}
\end{figure}

Figure \ref{fig:weight_ablation} shows the effect of increasing the weighting of the sparseBM loss compared to the ComboLoss. The validation Betti Matching error reduces from $>2500$ to $<250$ with increasing weight of the topological error. Validation dice scores are within $0.01$ for $\lambda<10^{-4}$ and start reducing notably ($>0.03$) for $\lambda>10^{-3}$.

\begin{figure}[h]
    \centering
    \includegraphics[width=\linewidth]{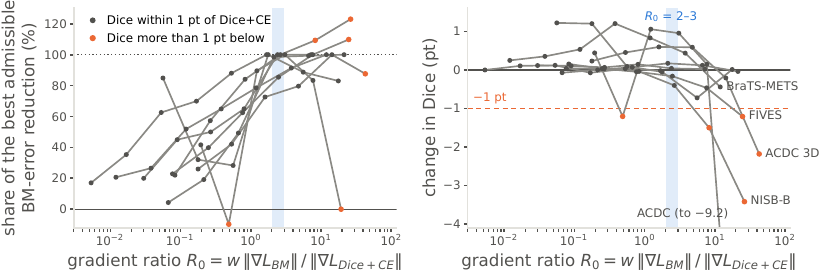}
    \caption{We observe that sparseBM's optimal weight can be chosen based on the ratio between the gradient norm of sparseBM and ComboLoss.}
    \label{fig:weight_rule}
\end{figure}

This effect can be explained when looking at the ratio between gradient norms of the loss components $R_0$ (\Cref{fig:weight_rule}). With increasing topology weight, sparseBM's gradient dominates the gradient of the ComboLoss, and the Betti Matching error decreases (left panel), while Dice performance remains stable. Once the ratio exceeds a threshold of $2-3$, the Dice performance starts to deteriorate (right panel).

\FloatBarrier
\subsection{Ablation on tau and network output distribution}
\label{sec:sup_tau_and_intensities}

\begin{figure}[h]
    \centering
    \includegraphics[]{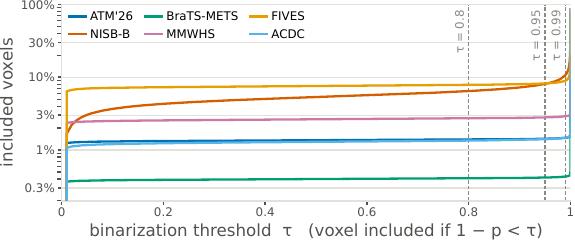}
    \caption{Fraction of included voxels as the sparsification threshold $\tau$ increases. The y-axis is logarithmic.}
    \label{fig:kept_vs_threshold}
\end{figure}

Table \ref{tab:tau} reports the influence of changing $\tau$ for the calculation of the sparseBM loss. Across the $\tau$ range of $0.5$--$0.9$, the kept fraction varies by only $0.02\%$; consequently, loss computation is comparable across these values of $\tau$. \Cref{fig:kept_vs_threshold} visualizes how the kept fraction is nearly constant across most of the sparsification-threshold range, with a small jump at the beginning (confident foreground) and a large jump near the background end of the filtration.

\begin{table}[t]
\centering
\scriptsize
\setlength{\tabcolsep}{4pt}
\caption{Kept fraction, runtime, validation BM error and Dice with varying $\tau$ on ATM'26 (validation). Bold: best $\tau$; the runtime differences between $\tau$ values are within 2\% (paired per micro-batch).}
\label{tab:tau}
\begin{tabular}{lrrrr}
\toprule
$\tau$ & kept (\%) & runtime (ms)$\downarrow$ & BM err.$\downarrow$ & Dice$\uparrow$ \\
\midrule
0.5 & 1.26 & 48.6{\tiny$\pm$6.2} & 4.40 & .9724 \\
0.6 & 1.26 & 49.1{\tiny$\pm$7.0} & 4.06 & .9734 \\
0.7 & 1.26 & 49.1{\tiny$\pm$7.1} & 4.09 & \textbf{.9740} \\
0.8 (default) & 1.27 & 49.3{\tiny$\pm$6.7} & 3.96 & .9732 \\
0.9 & 1.28 & 49.1{\tiny$\pm$6.7} & 4.14 & .9720 \\
0.95 & 1.29 & 50.0{\tiny$\pm$7.4} & \textbf{3.92} & .9733 \\
0.99 & 1.32 & 49.6{\tiny$\pm$6.2} & 3.99 & .9716 \\
\bottomrule
\end{tabular}
\end{table}

\FloatBarrier
\subsection{Details about the Experimental design}
\label{sec:app_experimental_design}

\subsubsection{Preprocessing}
\label{sec:app_preprocessing}
Following \citet{berger2025pitfalls}, we evaluate all datasets for the prevalence of topological artifacts and susceptibility to connectivity choices. In susceptible datasets (i.e., datasets where the global and local topology differ with different connectivity choices), we preprocess the label by adding single pixels to achieve well-composedness. The fill only adds voxels and changes little (e.g.\ a median of $0.16\,\%$ of the foreground on ATM'26 and a mean of $0.79\,\%$ on BraTS-METS). NISB-B is well-composed by construction. Our method (and its metrics) are based on the V-construction and therefore, implicitly following 6-connectivity (i.e., direct) for the foreground.

\subsubsection{Datasets.}
\label{sec:app_datasets}
We use six binary segmentation tasks, four in 3D and two in 2D, whose targets cover the topological regimes relevant for PH-based losses: a tree (ATM'26), thin sheets whose topology lives in the background (NISB-B), many small blobs whose number is the clinical quantity (BraTS-METS), a shell (MMWHS), a loopy 2D network (FIVES) and a ring (ACDC). \Cref{tab:datasets} summarizes them. Each dataset has a held-out test set that is not used for any design decision: loss weights are selected on the validation fold (fold 0, seed 42), and the selected configuration is trained with three seeds and scored once on the test set. 3D test volumes are predicted with sliding-window inference over the whole volume; 2D images are predicted in a single pass.

\begin{table}[h]
\centering
\scriptsize
\setlength{\tabcolsep}{3.5pt}
\caption{Datasets. Train/val = fold 0 of the cross-validation split; test = held-out cases, scored once.}
\label{tab:datasets}
\begin{tabular}{llllll}
\toprule
dataset & modality & target & train / val / test & resolution & patch \\
\midrule
ATM'26 & chest CT & airway tree & 193 / 50 / 53 volumes & native, 0.51--0.92\,mm in-plane & $128^3$ \\
NISB-B & synthetic EM & neuron boundaries & 5 / 1 / 1 volumes$^{a}$ & $4.5\times4.5\times10$\,nm (lifted) & $128^2\times64$ \\
BraTS-METS & brain MRI & metastases (tumour core) & 733 / 181 / 162 cases & 1\,mm isotropic & $128^3$ \\
MMWHS & cardiac CT & LV myocardium & 12 / 4 / 4 volumes$^{b}$ & 1.25\,mm isotropic & $128^3$ \\
\midrule
FIVES & fundus photography & retinal vessels & 478 / 120 / 200 images & native & $2048^2$ \\
ACDC & cine MRI & LV myocardium & 1506 / 396 / 1076 slices & native, 1.37--1.92\,mm & $224^2$ \\
\bottomrule
\multicolumn{6}{l}{$^{a}$ 4500 / 900 subvolumes for training / validation; the test set is 12 whole subvolumes ($601\times601\times301$ lifted voxels each).} \\
\multicolumn{6}{l}{$^{b}$ after selection on fold 0, all four folds are trained; see text.} \\
\end{tabular}
\end{table}

\paragraph{ATM'26.} The training release of Track 1 (binary airway segmentation) of the ATM'26 challenge \citep{zhang2023multi}
contains 299 chest CT volumes. We exclude the three volumes with slice spacing above twice the in-plane spacing and keep the native resolution, since resampling changes the Betti numbers of the thin airway labels. Of the remaining 296 volumes, 53 are held out for testing (stratified by voxel spacing) and 243 form five folds. The raw labels are a single tree under 26-connectivity but split into up to 144 components under 6-connectivity; the well-composedness fill reconnects them into one component on every volume.

\paragraph{NISB-B.} NISB \citep{rieger2024nisb}
provides synthetic electron-microscopy volumes with dense neuron instance labels. We segment the boundaries between objects, where the extracellular space counts as one additional object. The label is built on the interpixel grid: every voxel is lifted to a $2\times2\times2$ block, and a lifted voxel is foreground if the original voxels it lies between belong to different objects. This label is well-composed by construction, and the objects of interest are the components of its background. We use seven volumes of $3001\times3001\times1351$ voxels ($9\times9\times20$\,nm): five for training, one for validation and one for testing. Each volume is tiled into 900 overlapping subvolumes.

\paragraph{BraTS-METS.} We use the 2025 training release of BraTS-METS \citep{maleki2025analysis}
(1295 cases from 810 patients) and segment the tumour core (non-enhancing tumour core $\cup$ enhancing tumour), so that $\beta_0$ is the number of metastases. We keep the 1119 cases with anisotropy at most 2 and resample them to 1\,mm isotropic with a label-preserving scheme that keeps every lesion (plain nearest-neighbour resampling erases 63 of them). The 43 cases without tumour core are excluded. We hold out 108 patients (162 cases) for testing.

\paragraph{MMWHS.} We use the 20 labelled CT volumes of MM-WHS \citep{Zhuang2016MSMMA}
(the official test volumes have no public labels) and segment the left-ventricular myocardium, resampled to 1.25\,mm isotropic. Four volumes are held out for testing, and the remaining 16 form four folds of 12/4. Loss weights are selected on fold 0, as on the other datasets. Because the validation and test sets are small, each selected configuration is then trained on all four folds with three seeds each, and the reported numbers average all 12 models on the four test volumes.

\paragraph{FIVES.} FIVES \citep{jin2022fives}
contains 800 colour fundus images of $2048\times2048$ pixels with retinal vessel labels, split by the authors into 600 training and 200 test images from four diagnostic groups (AMD, diabetic retinopathy, glaucoma, normal). We exclude the two training images whose labels are empty, split the remaining 598 into five folds, and use the official 200 test images. Its labels are rich in the top homological dimension (on average 34 loops per image, against a single ring per ACDC slice). Networks take the full RGB image as input.

\paragraph{ACDC.} ACDC \citep{bernard2018deep}
provides cine MRI at end-diastole and end-systole for 100 training and 50 test patients. We segment the left-ventricular myocardium slice by slice (in-plane $224\times224$ at the native 1.37--1.92\,mm): padding slices are dropped, and real slices without myocardium are kept. In 2D the target is a single ring, avoiding the through-plane artefacts of the 5--10\,mm slice spacing in 3D. The five folds are grouped by patient (fold 0: 1506/396 slices), and the test set is the 1076 slices of the 50 official test patients.

\subsubsection{Pretraining, topological finetuning, and model selection}
Although sparse cubical complexes significantly reduce the runtime of PH-based methods, training with them still incurs overhead. In order to be able to conduct realistic experimentation as described above, we train in two phases: pre- and post-training. During pretraining, we follow the described, fixed procedure with ComboLoss until validation performance converges. We refer to this as the \emph{parent run}. Then, for each method, we initialize the model from this checkpoint and conduct a hyperparameter search, where each run is trained until convergence with the respective method. Additionally, we continue the parent run for the same number of epochs as a fair baseline. This approach is conducted for each dataset separately.

Each hyperparameter search varies the weight parameter $\lambda$, combining the respective loss function with the ComboLoss, and contains six runs per method. The run with the lowest Betti Matching error in validation that matches the parent run's Dice performance by $1pp$ is used for cross-seed experiments. There, we retrain each network with the chosen hyperparameter setting across three different seeds.

\begin{figure}[h]
    \centering
    \includegraphics[]{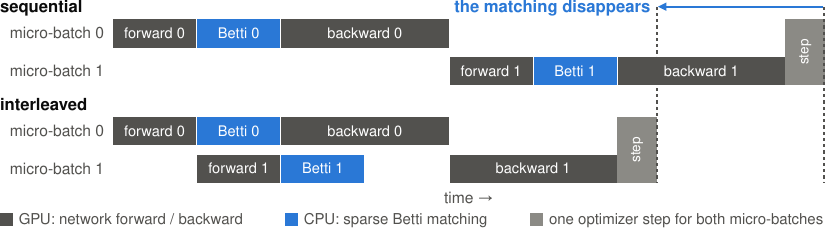}
    \caption{We interleave forward/backward passes with the loss calculation of two subsequent micro-batches to maximize GPU utilization and reduce runtime.}
    \label{fig:interleaved_training}
\end{figure}

\subsubsection{Interleaved training.}To further reduce runtime costs, we apply a further general optimization. First, we interleave two micro-batches by calculating the loss and doing the forward/backward pass on the GPU in parallel (see \cref{fig:interleaved_training}). This approach hides CPU runtime costs behind GPU calculations, which is typically the bottleneck in model training. From an optimization perspective, this approach is equivalent to gradient accumulation, which is a technique commonly used for training with larger effective batch sizes when GPU memory is limited. In our experiments, all methods for one dataset are trained with the same effective batch size.

\FloatBarrier
\subsection{Other applications}
\label{sec:app_other}
\subsubsection{SparseBM as segmentation metric}
\label{sec:app_metric}
\begin{figure}[h]
    \centering
    \includegraphics[]{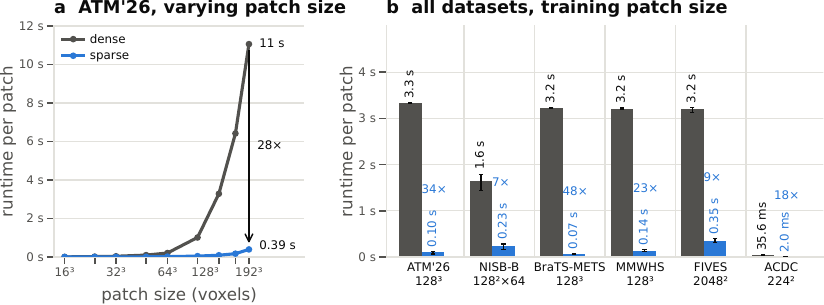}
    \caption{Runtime comparison for metric calculation between sparse and dense betti matching error. The runtime decrease translates to application on binary inputs.}
    \label{fig:metric_runtime}
\end{figure}

Sparse cubical complexes can be used for efficiently computing PH-based metrics, such as the Betti Matching error \citep{stucki2023topologically} or the Betti error \citep{hu2019topology}. As explained in \Cref{sec:method_segmentation}, the sparse metric is equivalent to its dense counterpart when computed between two binary inputs. We ran runtime experiments similar to our loss-function experiments and present the results in \Cref{fig:metric_runtime}. Similar to the loss function, we observe a substantial runtime reduction that increases with patch size and is most pronounced on the BraTS-METS dataset, with a $48$-fold reduction. This speedup enables efficient validation and topology-aware model selection, both of which are prohibitively expensive with dense PH-based approaches.

\subsubsection{Sparse cubical filtrations as a post-processing tool}
\label{sec:app_postprocessing}

In addition to their use as a segmentation loss, we evaluate sparse cubical filtrations as a post-processing tool in the ATM'26 airway challenge, where our method is ranked 3rd in the final test leaderboard. The goal of the challenge is to  create binary airway segmentations for chest CT scans. Submissions are ranked by the mean rank over Dice, clDice, tree-length detected (TLD), and branch detected (BD), where the latter two are measured on the largest connected component of the prediction. A distal branch that the network draws correctly but leaves disconnected from the main tree, therefore, counts as missed. We use the sparse complex to repair these failures. Every disconnected branch is a dimension-$0$ feature that dies when it merges with the main tree, and we reconnect it by raising the probability at its death voxels above the segmentation threshold.

\paragraph{Method.} Let $p$ be the predicted foreground probability and $g = 1 - p$. We binarize at $F_t = \{p > t\}$ and build the same sparse cubical complex that the loss uses, the V-construction on the kept voxels $\{g < \tau\}$, in dimension $0$ only and with $\tau=0.999$, so that every voxel with $p > 10^{-3}$ may carry a connection. Dimension-$0$ persistence is then a union--find sweep over the minimum spanning forest of the kept graph (edge weight $\max(g_u, g_v)$, elder rule), and every component whose bar dies below $\tau$ is a branch that the complex can still reach. For each of them, we raise the whole critical set of its death \citep{nigmetov2024big} --- the spanning-forest path between the two merging classes' birth vertices through the death edge --- above the threshold, which is the smallest set of voxels whose change realizes the merge. Merges are applied in order of death; components that remain essential are left alone. The largest $26$-connected component of the result is returned, so the output is connected by construction.

\paragraph{Results.} The kept set is a median of $0.33\%$ of the voxels. On the largest volume ($512\times463\times513$, $145$ components), the complete step takes $5.3$ seconds. The repairs are small and targeted with a median of $7$ added voxels per volume and at most $125$ voxels.

Measured against thresholding at $0.5$ followed by largest-component filtering, reconnection at $t = 0.1$ and $\tau = 0.999$ improves both tree metrics for every training loss, on $53$ held-out volumes and three seeds. The small Dice reduction comes from the lower binarization threshold and not from the added bridge voxels.

\begin{table}[h]
\centering
\caption{Results on the use of sparse cubical complexes as a post-processing tool. Each cell describes the improvements after post-processing compared to the respective base model.}
\begin{tabular}{lrrrr}
\toprule
loss & $\Delta$Dice & $\Delta$clDice & $\Delta$TLD & $\Delta$BD \\
\midrule
Dice+CE          & $-0.53$ & $-1.51$ & $+2.75$ & $+4.32$ \\
clDice           & $-0.70$ & $-1.13$ & $+2.55$ & $+4.15$ \\
Skeleton Recall  & $-0.71$ & $-1.68$ & $+2.20$ & $+3.51$ \\
\bottomrule
\end{tabular}
\label{tab:postprocessing}
\end{table}

\end{document}